\documentclass[lettersize,journal]{IEEEtran}
\usepackage{amsmath,amsfonts}
\usepackage{booktabs}
\usepackage{algorithmic}
\usepackage{array}
\usepackage[caption=false,font=normalsize,labelfont=sf,textfont=sf]{subfig}
\usepackage{textcomp}
\usepackage{stfloats}
\usepackage{url}
\usepackage{verbatim}
\usepackage{graphicx}
\usepackage{multirow}
\def\BibTeX{{\rm B\kern-.05em{\sc i\kern-.025em b}\kern-.08em
    T\kern-.1667em\lower.7ex\hbox{E}\kern-.125emX}}
\usepackage{balance}
\begin{document}
\title{CorMulT: A Semi-supervised Modality Correlation-aware Multimodal Transformer for Sentiment Analysis}


\author{Yangmin Li, Ruiqi Zhu, Wengen Li*%
\thanks{* Corresponding author.}%
\thanks{Yangmin Li is with Information Networking Institute at Carnegie Mellon University (CMU).}%
\thanks{Email: yangmin@cmu.edu}%
\thanks{Ruiqi Zhu is with School of Computer Science and Technology at Tongji University.}%
\thanks{Email: zhurq@tongji.edu.cn}%
\thanks{Wengen Li is with School of Computer Science and Technology at Tongji University.}
\thanks{Email: lwengen@tongji.edu.cn}
}

\markboth{IEEE Transactions on Affective Computing, Vol. XX, No. XX, Month Year}%
{Li et al.: CorMulT: Multimodal Sentiment Analysis}

\maketitle

\begin{abstract}
Multimodal sentiment analysis is an active research area that combines multiple data modalities, e.g., text, image and audio, to analyze human emotions, and benefits a variety of applications. Existing multimodal sentiment analysis methods can be roughly classified as modality interaction-based methods, modality transformation-based methods and modality similarity-based methods. However, most of these methods highly rely on the strong correlations between modalities, and cannot fully uncover and utilize the correlations between modalities to enhance sentiment analysis. Therefore, these methods usually achieve unsatisfactory performance for identifying the sentiment of multimodal data with weak correlations. To address this issue, we proposed a two-stage semi-supervised model termed Correlation-aware Multimodal Transformer (CorMulT) which consists of pre-training stage and prediction stage. At the pre-training stage, a modality correlation contrastive learning module is designed to efficiently learn modality correlation coefficients between different modalities. At the prediction stage, the learned correlation coefficients are fused with modality representations to make the sentiment prediction. According to the experiments on the popular multimodal dataset CMU-MOSEI, CorMulT obviously surpasses the state-of-the-art multimodal sentiment analysis methods. The code of CorMulT is available at 
https://github.com/YAMY1234/CorMulT.
\end{abstract}

\begin{IEEEkeywords}
Multimodal sentiment analysis, correlation analysis, semi-supervised learning, transformer
\end{IEEEkeywords}

\section{Introduction}\label{sec:introduction}
\IEEEPARstart{M}{ultimodal} sentiment analysis combines multiple perceptual modalities such as text, image, and audio to analyze human emotions. Currently, multimodal sentiment analysis has been widely applied in various domains, including predicting movie box office performance~\cite{Asur1}, predicting stock market performance~\cite{Bollen1}, and predicting political election results~\cite{Tumasjan1}. 

The general paradigm of multimodal sentiment analysis is to extract emotional information from multiple modalities and fuse them together to achieve a comprehensive and accurate understanding of emotions.
Existing methods for multimodal sentiment analysis can be roughly classified into three categories, i.e., \textit{modality interaction-based methods}, \textit{modality transformation-based methods}, and \textit{modality similarity-based methods}. Modality interaction-based methods emphasize on learning the interactions and dependencies between different modalities to combine them together. The representative methods are MultiModal InfoMax~\cite{Han1} and MMLatch~\cite{Paraskevopoulos1}. Modality transformation-based methods select one modality as the primary modality, and extract information from other modalities to obtain cross-modal representation. The representative methods include BAFN~\cite{Tang1} and RAVEN~\cite{Wang1}. Modality similarity-based methods compare the features of different modalities for deep understanding of sentiment, and the representative methods include Misa~\cite{Hazarika1} and MUET~\cite{liu2022umt}.

\begin{figure}[!t]
  \centering
  \includegraphics[width=\linewidth, height=3.5cm]{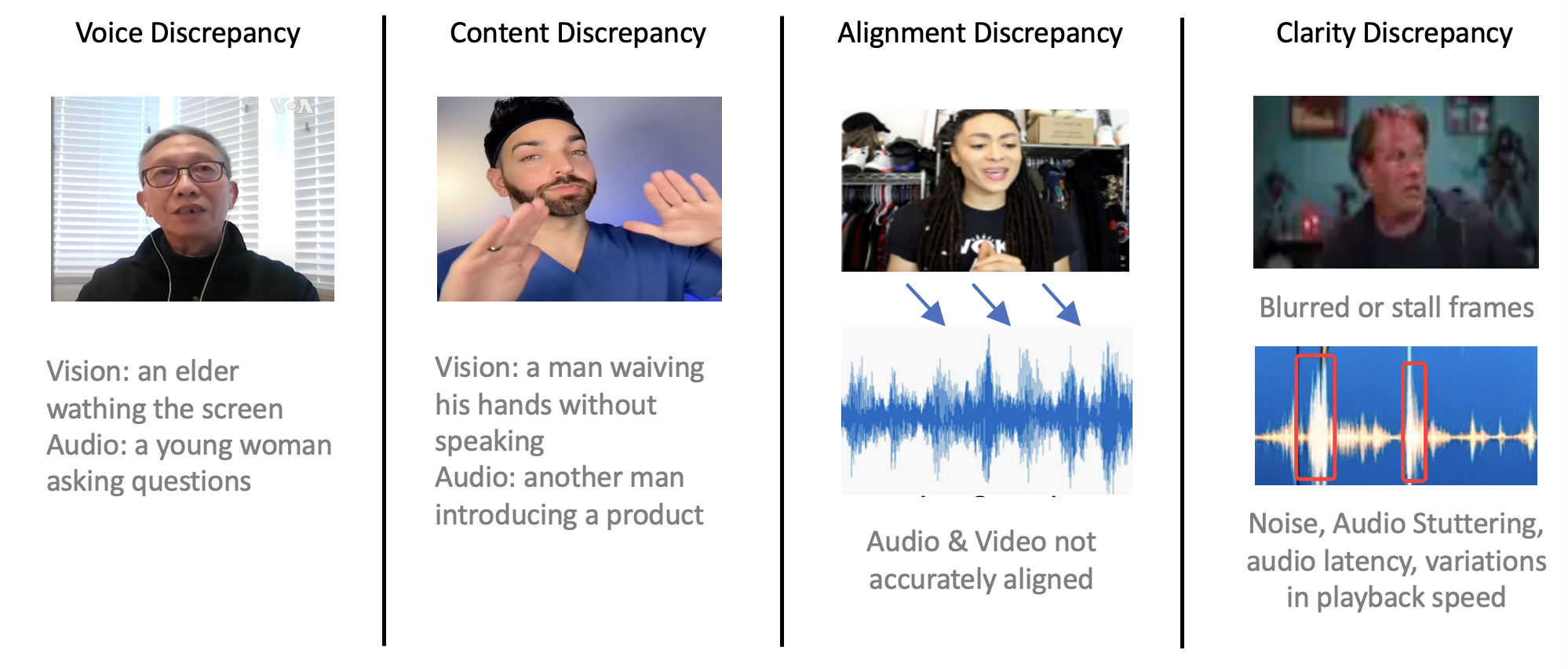}
  \caption{The multimodal samples with weak modality correlations, including voice discrepancy, content discrepancy, alignment discrepancy, and clarity discrepancy.}
  \label{fig:weak_corr_example}
\end{figure}

However, most existing methods for multimoal sentiment analysis heavily rely on the assumption of strong correlations between modalities, and have not yet fully explored and examined the correlations between modalites. In this case, they may achieve unsatisfactory performance for identifying the sentiment for multimodal data with weak correlations. In general, strong modality correlations are characterized by a well-aligned, consistent, and strongly related representation of objects across visual, textual, and audio aspects, with minimal noise interference such as irrelevant visual elements and background sounds. Conversely, weak modality correlations are often plagued with different discrepancies as illustrated in Fig.~\ref{fig:weak_corr_example}. Voice discrepancy arises when non-primary sounds overshadow the main audio subjects. Content discrepancy occurs when there is an inconsistency between the objects described in text and those depicted in images. Alignment discrepancy is the inability to achieve a complete synchronization across modalities. Clarity discrepancy refers to the presence of excessive noises that muddle the primary signals. In practice, it is difficult for existing multimodal sentiment analysis methods to accurately decide the sentiment of data samples with such discrepancies.

To address the above issue, this work proposed a two-stage semi-supervised model termed Correlation-aware Multimodal Transformer (CorMulT). CorMulT develops a novel modality correlation constractive learning module to learn modality correlation coefficients between different modalities at the pre-training stage. Then, the learned correlation coefficients are fused with the modality representations to produce the sentiment classification results. 

In sum, the contributions of this work are threefold as highlighted below.
\begin{itemize}
    \item We identified the issue of weak modality correlations in multimodal sentiment analysis, and proposed the CorMulT model to address this issue by accurately learning the correlations between modalities and fusing the learned correlations with modality representation to enhance sentiment analysis.
    \item We proposed the modality correlation evaluator, a pre-trained model based on contrastive learning, to learn the correlations between modalities by transforming multimodal features into a shared correlation space in which the distance between different modalities could be effectively quantified.
    \item We conducted extensive experiments to evaluate the effectiveness of the proposed method, and the results show that our method largely outperforms state-of-the-art multimodal sentiment analysis methods.
\end{itemize}

The remainder of the work is organized as follows. Section~\ref{sec:relatedwork} reviews the related work on multimodal sentiment analysis. Section~\ref{sec:methodology} presents the proposed CorMulT method in detail. Section~\ref{sec:experiments} reports the experimental results, and  Section~\ref{sec:conclusion} concludes this work.

\section{Related Work}\label{sec:relatedwork}
In this section, we provide more details about the three categories of multimodal sentiment analysis methods, i.e., \textit{modality interaction-based methods}, \textit{modality transformation-based methods}, and \textit{modality similarity-based methods}.

\subsection{Modality Interaction-based Methods}
These methods focus on capturing the dynamic interactions and dependencies between different modalities when combining information from multiple modalities. 
Wei et al.~\cite{Han1} proposed MultiModal InfoMax (MMIM) to maximize the mutual information between unimodal pairs, and that between the fusion results and low-level unimodal representations to preserve task-relevant information during multimodal fusion. 
Paraskevopoulos et al.~\cite{Paraskevopoulos1} utilized a forward feedback mechanism to capture the top-down cross-modal interactions and achieved multimodal fusion by masking sensory inputs with higher-level representations extracted from each modality. Kumar et al.~\cite{Kumar1} utilized self-attention to capture the long-term context and employed a gating mechanism to selectively learn weak contributing features. 

Sun et al.~\cite{sun2023general} proposed a general debiasing framework based on Inverse Probability Weighting (IPW), which adaptively assigns small weights to the samples with larger bias to avoid fitting the spurious correlations between multimodal features and sentiment labels. 
Xue et al.~\cite{9723597} proposed a multi-level attention map network (MAMN) to filter noise before multimodal fusion and capture the consistent and heterogeneous correlations among multi-granularity features. Quan et al.~\cite{9945514} proposed a multimodal comparative learning interaction module to better focus on the characteristics of different modalities. He et al.~\cite{10124248} proposed a multimodal mutual attention-based sentiment analysis (MMSA) framework that uses three levels of subtasks to preserve the unimodal unique semantics and enhance the common semantics of multimodal data. Wang et al.~\cite{s23052679} introduced supervised contrastive learning to learn effective and comprehensive multimodal data representation. Liu et al.~\cite{LIU2023110467} proposed the Scanning, Attention and Reasoning (SAR) model for multimodal sentiment analysis, where a perceptual scanning model is designed to perceive the image and text content, and the intrinsic correlations between them. Wang et al.~\cite{electronics12132986} proposed a model to assist in learning modality representations with multitask learning and contrastive learning. 

Tsai et al.~\cite{MULT} repeatedly reinforce the features of the target modality using low-level features of the source modality, thus ensuring cross-modal interactions and addressing the alignment issue between different modalities. Yu et al.~\cite{Yu1} proposed a cyclic memory enhancement network for capturing the long-term dependencies in multimodal data. Ping et al.~\cite{Ping1} proposed a novel temporal position prediction task for speech-text alignment to deal with the cross-modal alignment issue. Cheng et al.~\cite{10097669} and Ye et al.~\cite{10.1145/3551876.3554811} proposed Attentional Temporal Convolutional Network (ATCN) and Multimodal Temporal Attention (MMTA) to extract unimodal temporal features for adaptive inter-modal balance. Lin et al.~\cite{10143237} and He et al.~\cite{10.1117/12.2680996} narrowed the gap among different modalities by transferring the sentiment-related knowledge.

\subsection{Modality Transformation-based Methods}
In this category of methods, one modality serves as the primary modality while the others serve as supplementary modalities. For example, Khan et al.~\cite{Khan1} developed a two-stream model that converts image modality to textual representations as auxiliary text, thereby providing more information for the text modality. Wang et al.~\cite{Wang1} obtained the text drift vectors using non-textual data, e.g., audio and image modalities, thus achieving drifted word embedding. Majumder et al.~\cite{Majumder1} mapped the personality traits from textual documents to audio and visual modalities to improve the effectiveness of sentiment analysis. Tang et al.~\cite{Tang1} introduced a more discriminating text modality to guide the dynamic learning of emotional context within modalities, thus reducing the redundant information in auxiliary modalities during the modality transformation process. He et al.~\cite{9667257} fused non-verbal embedding with language embedding before inputting them into Bert to obtain multimodal representations. However, according to Gan et al.~\cite{Gan1}, pure modality transformation methods often underperform the methods that consider the interaction between modalities because modality transformation methods may overlook the inherent correlations between modalities and result in information loss.

\subsection{{Modality Similarity-based Methods}}
These methods compare the features across different modalities to facilitate the analysis of their sentiment. For instance, Yu et al.~\cite{Yu2} jointly trained multimodal and unimodal tasks to learn the consistency and difference in modality representations. Hazarika et al.~\cite{Hazarika1} projected each modality into two subspaces, i.e., modality-invariant subspace and modality-specific subspace, where modality-invariant subspace facilitates cross-modal representation learning to reduce modality discrepancies, and modality-specific subspace captures the distinctive features of each modality. Mai et al.~\cite{mai2022curriculum} proposed curriculum learning for weakly supervised modality correlations, and leveraged limited or noisy labeled data to align features across different modalities, thus learning a more discriminative embedding space for multimodal data. However, they often struggle with the ambiguity and sparsity of weak labels, which may lead to suboptimal learning of true modality correlations. In summary, although these methods are capable of learning certain correlations between modalities, they often fall short in further integrating the learned correlations for sentiment analysis, and are thus unable to enhance the analysis performance.

\subsection{{Discussion}}
As described above, most existing multimodal sentiment analysis models rely on the assumption of strong correlations between modalities. 
However, as discussed in Section~\ref{sec:introduction}, a large amount of multimodal data may not exhibit strong correlations. As a result, the correlations between modalities have not been fully explored and utilized in most models. Besides, it is labor-intensive and impractical to manually filter all the data samples with low modality correlations, and existing methods like cross-modal alignment also cannot solve the problem. Therefore, this work proposed the two-stage model CorMulT. At the pre-training stage, CorMulT learns the modality correlation coefficients via a modality correlation contrastive learning module. At the prediction stage, the learned modality correlation coefficients are fused with the modality representations to enhance the sentiment prediction.

\section{Methodology}\label{sec:methodology}
\subsection{Overview}
\begin{figure*}[!t]
  \centering
  \includegraphics[width=\textwidth]{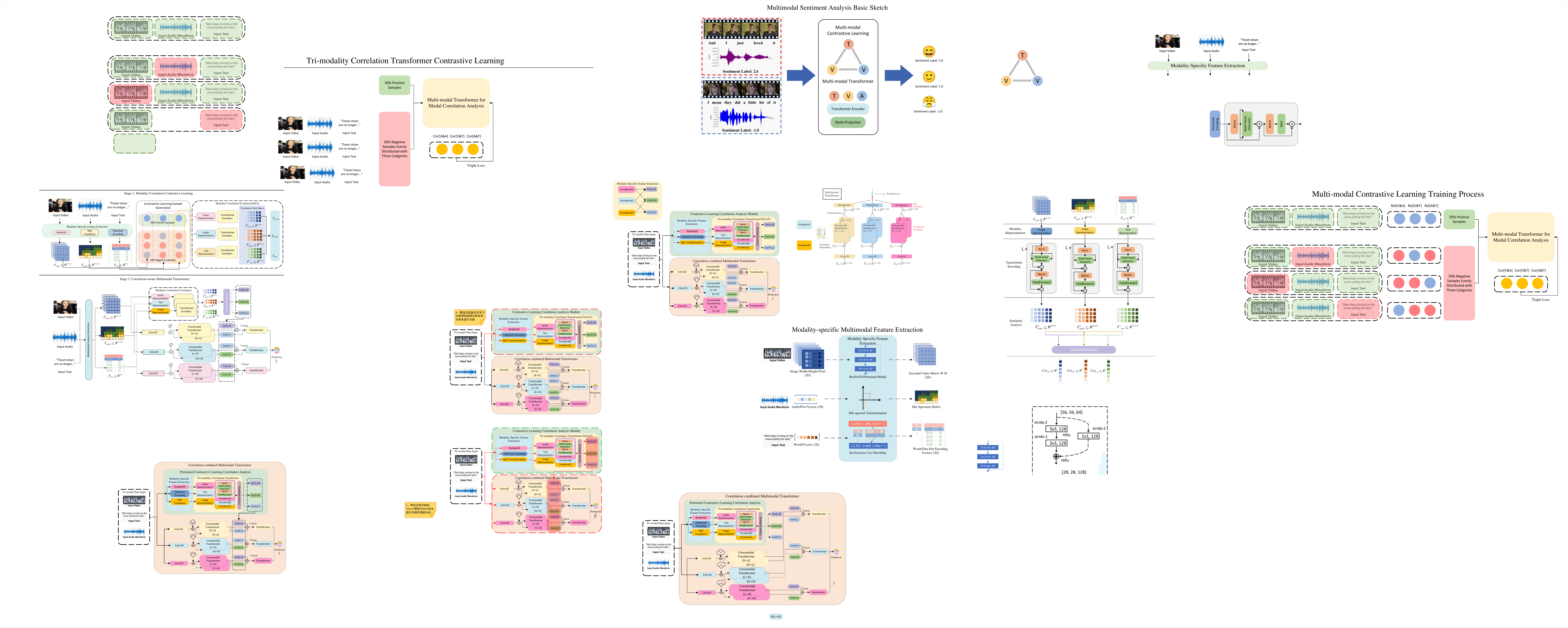}
  \caption{The framework of CorMulT Model which consists of modality correlation contrastive learning module and correlation-aware multimodal transformer.}


    \footnotesize{\textit{Note: The CorMulT model initiates with the Modality Correlation Evaluation (MCE) module using contrastive learning to determine modality correlation coefficients (Stage 1). The trained MCE is integral for enabling the Multimodal Transformer to understand and utilize the relationships between different modalities for more accurate predictions (Stage 2).}}
  \label{fig_overview}
\end{figure*}

Fig.~\ref{fig_overview} illustrates the architecture of Modality Correlation-aware Multimodal Transformer (CorMulT) which consists of modality correlation contrastive learning module and correlation-aware multimodal transformer module. Different modalities are first fed into the modality correlation contrastive learning module to obtain modality correlation coefficients which are then fused with the modality representations in the correlation-aware multimodal transformer module for predicting the sentiment.

\subsection{Modality Correlation Contrastive Learning}
Modality correlation contrastive learning module utilizes the contrastive learning framework to learn the correlation coefficients between different modalities. This module consists of two submodules, i.e., \textit{modality-specific feature extraction} and \textit{modality correlation evaluation}. The modality-specific feature extraction is responsible for extracting unimodal features from the three modalities, i.e., audio, image and text. The modality correlation evaluation module analyzes the correlations between different modalities.

\subsubsection{Modality-Specific Feature Extraction}
To encode different modalities into unimodal representations, different feature extraction methods are employed according to the characteristics of the modalities.

\textbf{For audio encoding}, we employ the Mel spectrogram~\cite{YANG202352} which is a widely used visual representation in audio processing that transforms the audio signals into an image-like format.
Initially, we capture the spectral characteristics of the audio signals using the Short-Time Fourier Transform (STFT) which decomposes the audio signals into  frequency components over short time intervals and produces a time-frequency representation, i.e.,
\begin{equation}
    X_{\text{stft}}(n, k) = \sum_{m=-\infty}^{\infty} x(m) w(m-n) e^{-j2\pi km/N}
\end{equation}
where \(X_{\text{stft}}(n, k)\) represents the coefficient value of the STFT matrix for the \(k^{th}\) frequency at the \(n^{th}\) frame,
\(x(m)\) denotes the value of input audio signal at time \(m\), \(w(m-n)\) denotes the value of window function at time \(n\), \(k\) denotes the sampling point on the frequency axis, and \(N\) represents the number of points in the Fast Fourier Transform (FFT).

Next, we convert the STFT representations into the Mel spectrogram which offers a more perceptually relevant representation of the audio. This conversion involves a set of Mel filters that mimic the human auditory system's response to different frequencies. These filters capture the energy within specific frequency ranges, i.e.,
\begin{equation}
H_m(k) =
\begin{cases}
    0 & \text{if } f_k < f_{m-1} \\
    \frac{f_k - f_{m-1}}{f_m - f_{m-1}} & \text{if } f_{m-1} \leq f_k < f_m \\
    \frac{f_{m+1} - f_k}{f_{m+1} - f_m} & \text{if } f_m \leq f_k < f_{m+1} \\
    0 & \text{otherwise}
\end{cases}
\end{equation}
where \(f_k\) represents the frequency value of the \(k^{th}\) sampling point on the frequency axis, \(f_m\) represents the center frequency of the \(m^{th}\) Mel frequency band, and \(H_m(k)\) denotes the output of the Mel filter bank for the \(k^{th}\) frequency. The Mel filter bank \(H_m(k)\) applies a set of triangular filters to map the STFT coefficients to the Mel scale. The dimension of \(H_m(k)\) is \(M \times K\), where \(M\) denotes the number of Mel frequency bands, and \(K\) denotes the number of STFT coefficients.

To further enhance the audio signal representation, we calculate the logarithm of the Mel spectrogram. This process transforms the power spectrum obtained from the STFT into a logarithmic scale, which more closely mirrors human auditory perception. The logarithmic Mel spectrogram is represented as follows:
\begin{equation}
    X_{\text{mel}}(n,m) = \log\left(\sum_{k=0}^{N-1} |X_{\text{stft}}(n,k)|^2 H_m(k)\right)
\end{equation}
where \(X_{\text{mel}}(n,m)\) denotes the log-transformed Mel spectrogram value for the \(m^{th}\) Mel frequency band at the \(n^{th}\) time frame. The Mel spectrogram matrix \(X_{\text{mel}}\) is an element of the set \( F_{A} \), which is a subset of \( \mathbb{R}^{b \times t \times m} \), where \(b\) represents the batch size, \(t\) denotes the number of time frames, and \(m\) represents the number of Mel frequency bands. This logarithmic scaling of the Mel spectrogram not only enhances the representation of the audio signals by emphasizing the perceptually important aspects of the sound, but also stabilizes the numerical range, thus facilitating more effective model training.

By employing the Mel spectrogram and its logarithm, we obtain a compact and informative visual representation of the audio signal, enabling our model to well learn the patterns in the frequency content of the audio.

\textbf{For text encoding}, the input text \( T \) is first tokenized using the BertTokenizer to generate a sequence of tokens \( W \). Next, a vocabulary \( V \) is created using the CMU\_MOSEI dataset, and \( V \) contains a collection of words and their corresponding numerical representations. 
The tokens \( W \) are then encoded using the vocabulary \( V \) to obtain the encoded sequence \( S \). 
To ensure uniformity in the input size, the encoded sequence \( S \) is padded or truncated to a pre-defined sequence length. 
The final representation of the text input is \( F_{T} \subseteq \mathbb{R}^{b \times s} \), where \( b \) is the batch size and \( s \) is the sequence length. 

\textbf{For image encoding}, 
we first decode the video data \( V \) into image frames using the OpenCV library. 
To extract high-level features from image frames, we utilize ResNet50 to learn the feature representation of each image frame \(I_k\).
To reduce the dimension of the extracted features, we employ pooling operation to aggregate information within local regions, thus capturing the most salient features while reducing the overall dimension. 
The final video representation \( F_{V}\) is obtained in an \( R^{b \times f \times i} \) space, where \( b \) represents the batch size, \( f \) corresponds to the number of image frames, and \( i \) is the feature dimension. 

\subsubsection{Sample Generation for Contrastive Learning}
Most existing methods for multimodal sentiment analysis heavily rely on the manually annotated datasets. However, few datasets have explicit annotation for modality correlations. Therefore, we introduce contrastive learning to unify the unimodal data representations. To this end, the critical task is to generate effective positive and negative data samples. 

Since most video data in our daily life exhibits inter-modality correlations, we thus employ multiple enhancement techniques, including \textit{temporal shift}, \textit{cross-sample mixing}, and \textit{data perturbation}, to generate negative samples. Temporal shift slightly shift the time alignment of different modalities within a sample to create negative samples. For example, for the audio and image modalities of a video, they can be shifted slightly out of sync to create negative samples.
Cross-sample mixing mix modalities across different samples to create negative samples. For example, the audio from one video can be mixed with the visual component of another video to create negative samples. Data perturbation introduces perturbations, e.g., noise and image transformations, to produce negative samples. For example, adding noise to an image can create a negative sample that is similar to the original image.

\subsubsection{Modality Correlation Evaluation}
\begin{figure*}[!t]
  \centering
  \includegraphics[width=0.65\linewidth]{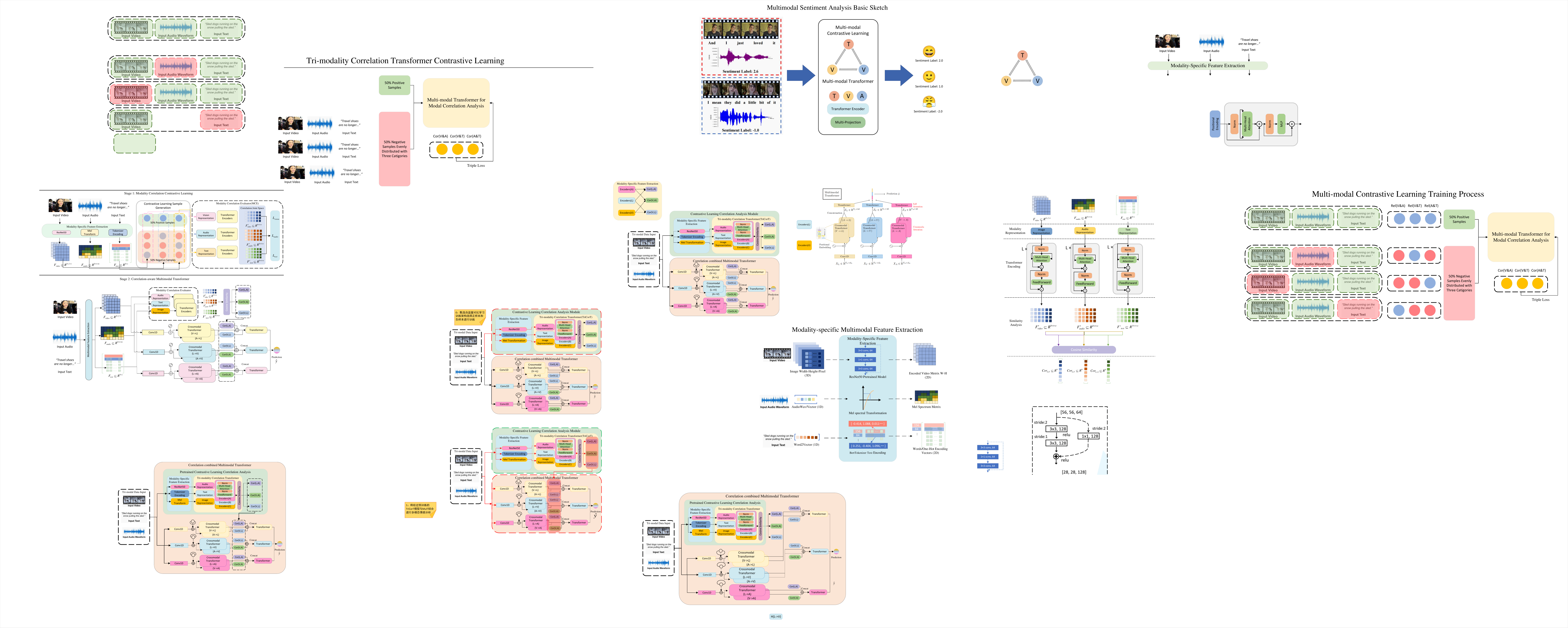}
  \caption{The structure of Modality Correlation Evaluation (MCE) module which learns the modality correlation and outputs the coefficients among modalities. }
  \label{fig_TriCorT}
\end{figure*}

The unimodal representations are then fed into modality correlation evaluation (MCE) module to evaluate the correlations between different modalities. 
Fig.~\ref{fig_TriCorT} shows the structure of MCE module.
which consists of modality transformation layer, cross-attention layer, feedforward layers, similarity analysis layer, and loss calculation layer.
The modality transformation layer transforms the input data into higher-level representations with dimensions equal to the hidden layer dimension. The cross-attention layer calculates the attention scores across modalities to reveal the interrelations among text, audio, and images. The feedforward layers linearly transform the attention results, i.e., mapping them to the pre-defined output dimension. The goal of MCE is to minimize the triplet loss function which calculates the disparity between the predicted outputs and labels. 

\textbf{Modality representation.} For audio modality, its unimodal representation \(F_{A} \in R^{b\times t\times m}\) is converted to a representation of dimension \(F_{A}^{\prime} \in R^{b\times t\times d}\) after positional encoding, where \(d\) is the hidden layer dimension.
For text modality, after the positional encoding and textual embedding, the unimodal representation \(F_{T} \in R^{b\times s}\) is converted to \(F_{T}^{\prime} \in R^{b\times s\times d}\), where \(s\) denotes the text sequence length.
For image modality, after positional encoding and visual embedding, its unimodal representation \(F_{V} \in R^{b\times f\times i}\) is transformed to \(F_{V}^{\prime} \in R^{b\times f\times d}\).


\textbf{Transformer encoding.} The encoder consists of three layers, and takes as inputs \(F_{A}^{\prime}, F_{T}^{\prime},\) and \(F_{V}^{\prime}\) to generate compact and consistent representation for each modality. The implementation details of the encoder are discussed as follows.

First, the input \(X\) is normalized to obtain \(X_1\) using the LayerNorm function, i.e., 
\begin{equation}
    X_1 = \text{LayerNorm}(X)
\end{equation}
where $X$ denotes the unimodal representation of certain modality. Next, \(X_1\) is passed through the Multihead Attention to produce \(X_2\), i.e.,
\begin{equation}
    X_2 = \text{MultiheadAttention}(Q,K,V:X_1)
\end{equation}
where \(X_1\) is used as query \(Q\), key \(K\), and value \(V\).

To well preserve the original information, a residual connection is established between \(X_2\) and $X_1$, and the results are normalized using LayerNorm, i.e.,
\begin{equation}
    X_3 =  \text{LayerNorm}(X_1 + X_2)
\end{equation}

The normalized \(X_3\) is then fed into the FeedForward layer, and the results are combined with \(X_3\) to produce the final output of tranformer encoding, i.e., 
\begin{equation}
    X_{out} = X_3 + \mathrm{FeedForward}(X_3)
\end{equation}

Transformer encoding effectively encodes the inputs \(F_{A}^{\prime}, F_{V}^{\prime},\) and \(F_{T}^{\prime}\) for different modalities to learn their compact representations for further processing.

\textbf{Similarity analysis.} After the features are encoded, linear transformation is implemented to project \(X_{out}\) for each modality to a shared correlation space $R^{b\times t\times o}$ for similarity comparison, where \(t\) is the temporal output mapping parameter and \(o\) specifies the dimensionality of features in the output space. Thus, we obtain \(F_{A}^{\prime\prime}, F_{V}^{\prime\prime},\) and \(F_{T}^{\prime\prime}\) for audio, vision, and text modalities, respectively, space $R^{b\times t\times o}$.
Then, we use cosine similarity to measure the correlations between different modalities, and obtain the pair-wised cross-modal correlation coefficients, i.e., 
${Cor}_{T,A}$, ${Cor}_{T,V}$ and ${Cor}_{V,A}$. 

\subsubsection{Loss Calculation.}




We designed a triple-loss function (\(\text{TL}\)) to minimize the disparity between each modality's feature matrix in space \(R^{b\times t\times o}\) and those of the other modalities, thus enhancing the interaction among modalities to improve joint representation learning.

Concretely, the \(\text{TL}\) function measures and optimizes the relationship between the feature matrix of each modality (\(F'\)) against positive (\(F_{p}\)) and negative (\(F_{n}\)) feature matrices of another modality, where "positive" refers to the highly correlated modalities, and "negative" indicates the uncorrelated ones. The aim is to foster closer feature vectors among related modalities while enlarging the distance between unrelated ones.

The \(\text{TL}\) function is defined as:
\begin{equation}
\text{TL}(F', F_p, F_n) = \max(0, \text{d}(F', F_p) - \text{d}(F', F_n) + \mu)
\end{equation}
where \(\text{d}\) represents a distance function and \(\text{margin}\) is a predefined threshold, and \(\mu\) represents the margin set as 0.2 in our experiment.

For three modalities, the corresponding losses, i.e., \(LOSS_A\), \(LOSS_T\), and \(LOSS_V\), are derived from their similarity with the features of the other two modalities:
\begin{equation}
\begin{split}
    LOSS_A &= \text{TL}(F_A', F_{T_p}, F_{T_n}) + \text{TL}(F_A', F_{V_p}, F_{V_n})  \\
    LOSS_T &= \text{TL}(F_T', F_{A_p}, F_{A_n}) + \text{TL}(F_T', F_{V_p}, F_{V_n}) \\
    LOSS_V &= \text{TL}(F_V', F_{T_p}, F_{T_n}) + \text{TL}(F_V', F_{A_p}, F_{A_n}) \\
\end{split}
\end{equation}

The final loss for modality correlation evaluation is \(LOSS = \frac{LOSS_A + LOSS_T + LOSS_V}{3}\), which could enhance the interactions among different modalities and ensure an equal contribution from each modality towards the overall learning objective.

\subsection{Correlation-aware Multimodal Transformer} 
Correlation-aware multimodal transformer receives as inputs the features extracted from the audio, image, and text modalities by the feature extraction module, and merges them with the modality correlations learned by the MCE module for sentiment analysis. 

\subsubsection{Multimodal Feature Learning}
For each modality \(m \in \{T, A, V\}\), where \(T\), \(A\), and \(V\) denote the text, audio, and visual data, respectively, we project the extracted feature $F_m$ via a 1D convolution layer to a common feature dimension to obtain \(F_m'\), i.e.,
\begin{equation}
F_m' = \text{Conv1D}(F_m)
\end{equation}

After projection, we apply two successive crossmodal transformer layers to align unmodal representations. Each layer takes the projected features of one modality as the query, and the features of another modality as the key and value to achieve crossmodal interactions, i.e.,
\begin{equation}
\begin{aligned}
h_{T \to V} &= \text{CrossmodalTransformer}_{T \to V}(F_T', F_V') \\
h_{T \to A} &= \text{CrossmodalTransformer}_{T \to A}(F_T', F_A') \\
h_{V \to T} &= \text{CrossmodalTransformer}_{V \to T}(F_V', F_T') \\
h_{V \to A} &= \text{CrossmodalTransformer}_{V \to A}(F_V', F_A') \\
h_{A \to T} &= \text{CrossmodalTransformer}_{A \to T}(F_A', F_T') \\
h_{A \to V} &= \text{CrossmodalTransformer}_{A \to V}(F_A', F_V')
\end{aligned}
\end{equation}
where \(h_{M \to N}\) represents the output of the crossmodal transformer layer, and modality \(M\) is enriched with the information from modality \(N\). 

\subsubsection{Modality Correlation-enhanced Multimodal Fusion}
Both inter and intra-modality information are incorporated with the results of modality correlation analysis. First, we perform correlation analysis using the pre-trained MCE model, i.e., 

\begin{equation}
\text{Cor}_{T,A}, \text{Cor}_{T,V}, \text{Cor}_{A,V} = \text{MCE}(F_A, F_T, F_V)
\end{equation}

where \( \text{Cor}_{T,A}\), \(\text{Cor}_{T,V}\) and \(\text{Cor}_{A,V} \) represent the pair-wised modality correlation coefficients.

Then, we obtain the feature representation \( h_{m_i \to m_j} \) for each pair of modality \( m_i, m_j \in \{T,A,V\} \) using the cross-modal attention and self-attention layers. These features are integrated with modality correlation coefficients to produce comprehensive representations that encompass both intra-modal and inter-modal information, i.e.,
\begin{equation}
\begin{aligned}
h_{T \to A}^{\prime} &= h_{T \to A} \times \text{Cor}_{T,A} \\
h_{T \to V}^{\prime} &= h_{T \to V} \times \text{Cor}_{T,V} \\
h_{A \to T}^{\prime} &= h_{A \to T} \times \text{Cor}_{T,A} \\
h_{A \to V}^{\prime} &= h_{A \to V} \times \text{Cor}_{A,V} \\
h_{V \to T}^{\prime} &= h_{V \to T} \times \text{Cor}_{T,V} \\
h_{V \to A}^{\prime} &= h_{V \to A} \times \text{Cor}_{A,V}
\end{aligned}
\end{equation}
where \( h_{T \to A}^{\prime}, h_{T \to V}^{\prime}, h_{A \to T}^{\prime}, h_{A \to V}^{\prime}, h_{V \to T}^{\prime}, h_{V \to A}^{\prime} \) are the modality-specific feature representations enhanced by their corresponding modality correlation coefficients, signifying the weighted importance of one modality's features in the context of another.

Subsequently, we utilize the self-attention to further refine the fused features for culminating in a holistic representation, i.e.,
\begin{equation}
\begin{aligned}
H_T' &= \text{Transformer}_{T_{mem}}(\text{Concat}(h_{A \to T}^{\prime}, h_{V \to T}^{\prime})) \\
H_A' &= \text{Transformer}_{A_{mem}}(\text{Concat}(h_{T \to A}^{\prime}, h_{V \to A}^{\prime})) \\
H_V' &= \text{Transformer}_{V_{mem}}(\text{Concat}(h_{T \to V}^{\prime}, h_{A \to V}^{\prime}))
\end{aligned}
\end{equation}
where \( H_T', H_A', H_V' \) are the final feature representations for text, audio and visual modalities, respectively.
Then, the learned representations for three modalities are concatenated together to obtain the comprehensive representation about the multimodal data sample, i.e.,
$H_{concat} = \text{Concat}(H_T', H_A', H_V')$

Finally, the sentiment prediction is attained through a linear projection layer and a Softmax layer. The most probable sentiment class is determined by the index of the maximum value in the Softmax output, i.e.,
\begin{equation}
y = \text{Softmax}(\text{Linear}(H_{concat}))
\end{equation}
where \( y \) is the probability distribution over sentiment classes, and the predicted sentiment class \( \hat{c} \) is obtained by:
\begin{equation}
\hat{c} = \arg \max_i (y_i)
\end{equation}
where \( \hat{c} \) is the index of the highest probability in vector \( y \), corresponding to the most likely sentiment class.

\section{Experiments}\label{sec:experiments}
\subsection{Dataset and Settings}

\subsubsection{Dataset}
We conduct experiments primarily on the CMU\_MOSEI dataset, a widely used benchmark for multimodal sentiment analysis, and supplement our study with additional validation on the CH-SIMS dataset to assess model generalizability across different datasets.

The CMU\_MOSEI dataset comprises 3,228 distinct video clips spanning various domains such as reviews, debates, and consultations. Each clip contains three modalities—video, audio, and text—and is annotated with seven fundamental emotion classes on an intensity scale ranging from -3 to 3. In total, the dataset includes 23,453 sentences, where each sentence, along with its corresponding audio and video segments, forms a sample. Table~\ref{tab:dataset} presents the statistical details of the CMU\_MOSEI dataset. For our experiments, we split the dataset into 70

To further evaluate the model's performance on a different dataset, we also conduct experiments on CH-SIMS, a newly established benchmark for multimodal sentiment analysis in Chinese. The CH-SIMS dataset consists of 2,281 curated video segments sourced from movies, TV series, and variety shows. Each segment is annotated with both multimodal and unimodal sentiment labels, facilitating a detailed analysis of modality-specific contributions. The dataset covers three modalities—text, visual, and audio—and sentiment labels are categorized into five levels: Negative, Weakly Negative, Neutral, Weakly Positive, and Positive. Table~\ref{tab:dataset} provides the detailed statistics of the CH-SIMS dataset.

This setup allows us to conduct a comprehensive set of comparisons and ablation studies on CMU\_MOSEI while verifying the model’s robustness and effectiveness on CH-SIMS.

\begin{table}[h]
\centering
\caption{The overview of CMU\_MOSEI dataset.}\label{tab:dataset}
\begin{tabular}{ll}
\toprule
\textbf{Feature} & \textbf{Description} \\
\midrule
Source & YouTube \\
Physical Size & 106GB \\
Total Samples & 23,453 \\
Video Clips & 3,228 \\
Frame Rate & 30Hz \\
Unique Speakers & 1,000 \\
Distinct Topics & 251 \\
Avg. Sentence Number & 7.3 \\
Avg. Sentence Length & 7.28 seconds \\
Modalities & Text (l), Visual (v), Audio (a) \\
Labels & Emotion [-3,3] \\
Total Duration & 65 hrs 53 mins 36 secs \\
Primary Domains & Reviews 16.2\%, Debates 2.9\%, Consults 1.8\% \\
\bottomrule
\end{tabular}
\end{table}

\begin{table}[h]
    \centering
    \caption{The overview of CH-SIMS dataset.}
    \label{tab:dataset}
    \begin{tabular}{ll}
        \toprule
        \textbf{Feature} & \textbf{Description} \\
        \midrule
        Source & Movies, TV series, Variety shows \\
        Total Samples & 2,281 \\
        Video Segments & 2,281 \\
        Frame Rate & 30Hz \\
        Unique Speakers & 474 \\
        Average Segment Length & 3.67 seconds \\
        Average Word Count & 15 words \\
        Modalities & Text (T), Visual (V), Audio (A) \\
        Sentiment Annotations & Multimodal and Independent Unimodal \\
        Labels & Three categories: Negative,Neutral, Positive \\
        \bottomrule
    \end{tabular}
\end{table}

\subsubsection{Experiment Settings}
All experiments were conducted on a server equipped with two Intel SkyLake-E 12-core processor and a GeForce GTX 2080 graphics card. We run each experiment five times, and report the average results. Multiple evaluation metrics as listed below are introduced to evaluate the performance of CorMult model and the baseline multimodal sentiment analysis methods.
\begin{itemize}
    \item \textbf{Acc@7}: the accuracy of predicting all seven emotion class labels.
    \item \textbf{Acc@2}: the accuracy of predicting the top two emotion class labels.
    \item \textbf{F1}: the harmonic mean of prediction precision and recall.
    \item \textbf{Corr}: the correlation coefficient between the prediction results and true emotion values.
\end{itemize}

\subsection{Effects of Perturbation Strategy}
The following four negative sample generation strategies are considered for contrastive learning.

\textbf{Strategy A:} For each data sample, we randomly choose one modality and replace it with the corresponding modality from another randomly selected data sample in the same batch.

\textbf{Strategy B:} For audio and video modalities, we introduce a time offset, e.g., one second, between them. For text, an offset of one word is introduced.

\textbf{Strategy C:} For audio and video modalities, a Gaussian noise is added. For text, words or characters are randomly replaced. 

\textbf{Strategy D:} A combination of strategies A, B, and C, i.e., randomly selecting one of the three strategies above.

Each strategy is applied across the entire training dataset. Training parameters and model architectures are consistent across strategies to ensure a fair comparison. 
Table~\ref{table:corMult_performance} presents the sentiment analysis results while using different data perturbation strategies.
Among the four strategies, Strategy D achieves the highest values in terms of all five performance evaluation metrics, indicating the advantage of combining multiple data perturbation strategies.


\begin{table}[h]
\centering
\caption{The results of CorMulT when using different data perturbation strategies for contrastive learning in MCE module.}
\label{table:corMult_performance}
\begin{tabular}{@{}lccccc@{}}
\toprule
Strategy & $Acc@7$ & $Acc@2$ & $F1$ & $Corr$ \\ 
\midrule
A & 52.1 & 85.9 & 85.7 & 0.747 \\
B & 51.2 & 83.1 & 83.7 & 0.759 \\
C & 52.1 & 83.3 & 83.9 & 0.754 \\
D & \textbf{52.8} & \textbf{86.5} & \textbf{86.7}  & \textbf{0.779} \\
\bottomrule
\end{tabular}
\end{table}

\subsection{Effects of Distance Measurement}
The MCE module in CorMult model maps the learned audio, image, and text to a unified shared space in which the semantic correlations between different data modalities can be well evaluated. We seek to identify the most appropriate distance metric to measure modality correlations. 
Concretely, given two vectors \(A\) and \(B\), the metrics under consideration include Euclidean distance, Manhattan distance, Chebyshev distance, cosine similarity, and Mahalanobis distance, i.e.,

(1) Euclidean Distance:
\begin{equation}
d(A, B) = \left\|A-B\right\|_2 = \sqrt{\sum_{i=1}^{n} (a_i-b_i)^2}
\end{equation}
    
(2) Manhattan Distance:
\begin{equation}
d(A, B) = \left\|A-B\right\|_1 = \sum_{i=1}^{n} \left|a_i-b_i\right|
\end{equation}
    
(3) Chebyshev Distance:
\begin{equation}
d(A, B) = \left\|A-B\right\|_\infty = \max_{i=1}^{n} \left|a_i-b_i\right|
\end{equation}
    
(4) Cosine Similarity:
\begin{equation}
\text{similarity}(A, B) = \frac{A \cdot B}{\left\|A\right\| \cdot \left\|B\right\|}
\end{equation}
    
(5) Mahalanobis Distance with covariance matrix \( C \):
\begin{equation}
d(A, B) = \sqrt{(A-B) \cdot C^{-1} \cdot (A-B)^T}
\end{equation}
where $d(A, B)$ denotes the distance between vectors $A$ and $B$, and \(a_i\) and \(b_i\) represent the \(i^{th}\) components of vectors \(A\) and \(B\), respectively. 




Table~\ref{table_distance} presents the results of CorMulT while using different distance measurement methods.
According to the results, cosine similarity emerges as the optimal distance metric for MCE, and could efficiently capture the inter-modal correlations.

\begin{table}[h!]
\centering
\caption{Results of CorMulT while using different distance measurement methods.}\label{table_distance}
\begin{tabular}{cccccc}
\toprule
\textbf{Distance} & \textbf{Acc@7} & \textbf{Acc@2} & \textbf{F1} & \textbf{Corr} \\
\midrule
Euclidean Distance & 52.6 & 86.3 & 86.4 & 77.4 \\
Manhattan Distance & 52.5 & 86.0 & 85.6 & 77.4 \\
Chebyshev Distance & 52.6 & 85.7 & 86.3 & 76.9 \\
Cosine Similarity & \textbf{52.8} & \textbf{86.5} & \textbf{86.7} & \textbf{77.9} \\
Mahalanobis Distance & 52.4 & 85.8 & 86.0 & 77.6 \\
\bottomrule
\end{tabular}
\end{table}

\subsection{Parameters Tuning}
\begin{figure*}[h]
  \centering
  \includegraphics[width=1\textwidth]{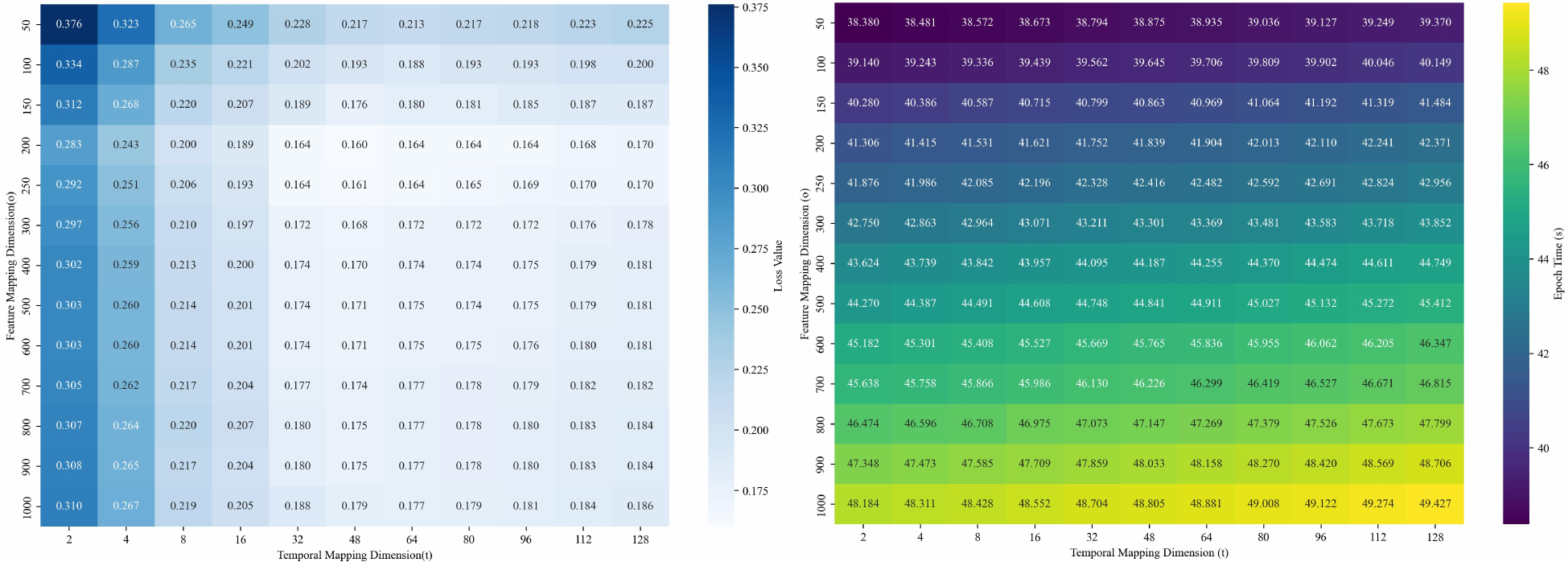}
  \caption{The heatmap of training loss and epoch time while varying temporal mapping dimension and feature mapping dimension.}
  \label{fig_para}
\end{figure*}

Temporal mapping dimension \( t \) and feature mapping dimension \( o \) are two important parameters in STAD model. We thus conduct experiments to find appropriate values for them using the grid search method. 
Fig.~\ref{fig_para} illustrates the training loss and epoch time(the training time for each epoch during the training of MCE) of STAD while varying \( t \) from 2 to 128, and varying \( o \) from 50 to 1000.
To mitigate the randomness, for each combination \( (t, o) \), we run experiments five times, and report the average results. Considering the large number of combinations, only the first 10\% of the CMU\_MOSEI dataset was used for tuning parameters, thus reducing the time cost.
According to Fig.~\ref{fig_para}, when \( t \) is between 2-48, higher losses were observed, but when  \( t \) is higher than 48, the loss maintained between 0.16-0.18. Best results appeared within \( t = 48-64 \), slightly increasing after 64. For \( o \), results were generally better when  \( o \) is higher than 200, with a slight decline in performance within the 300-1000 range. The optimal range for \( o \) was determined to be 200-300. As the epoch time is also increasing slightly as \( t \) and  \( o \) increase, we need to find the parameters that achieve the comparatively best results but should not introduce high latency. Hence, based on a combined assessment of loss and efficiency, the chosen dimensions for shared space are \( (64, 200) \).

\subsection{Model Comparison}
On CMU\_MOSEI dataset, CorMulT model is compared with multiple representative multimodal sentiment analysis models, including EF-LSTM\cite{williams-etal-2018-recognizing}, LMF\cite{LMF}, TFN\cite{TFN}, MFM\cite{Li2023MultimodalFM}, CM-Bert\cite{CM-Bert}, MISA\cite{Hazarika1}, BBFN\cite{Han1}, MULT\cite{MULT}, ICCN\cite{ICCN}, MMLatch\cite{Paraskevopoulos1}, TMSON\cite{TMSON}, DVMI\cite{DVMI}, ASBA-TLRF\cite{ABSA-TLRF}, EMT\cite{EMT} and MMIN\cite{MMIN}.
Table~\ref{tab:comparison} presents the results of CorMulT and baseline methods, where V+A+T indicates that the model utilizes vision, audio and text modalities. CM-Bert only uses vision and text modalities, i.e., V+T. Among all models, CorMulT exhibits superior performance.
Specifically, considering the F1 score and accuracy metrics (Acc@2 and Acc@7), CorMulT distinctly outperforms other models. For Acc@7, CorMulT achieves 52.8\%, surpassing the next best, i.e., MMLatch, at 52.1\%. Similarly, for Acc@2, CorMulT scores 86.8\%, surpassing the next best, i.e., DVMI, at 86.6\%.
Regarding the F1 score, CorMulT's 86.7\% markedly eclipsed others, showcasing its robustness since F1 is the harmonic mean of precision and recall.
Lastly, examining Corr (correlation), CorMulT excels with a score of 0.779, substantially ahead of EMT's 0.774, underscoring its adeptness in comprehending multimodal data intricacies. 

On CH-SMIS dataset, CorMulT model is compared with multiple representative multimodal sentiment analysis models, including EF-LSTM\cite{williams-etal-2018-recognizing}, LF-DNN\cite{socher-etal-2013-recursive}, TFN\cite{TFN}, LMF\cite{LMF}, MFM\cite{Li2023MultimodalFM}, Graph-MFN\cite{Graph-MFN}, MULT\cite{MULT}, MLF-DNN\cite{CH-SIMS}, MTFN\cite{CH-SIMS}, Self-MM\cite{Self-MM}, DVMI\cite{DVMI}, M\textsuperscript{3}SA\cite{M3SA} and EMT\cite{EMT}.
Table~\ref{tab:ch_sims_comparison} presents the results of CorMulT and baseline methods.
Specifically, considering the F1 score and accuracy metrics (Acc@2 and Acc@3), CorMulT distinctly outperforms most of other models. For Acc@2, CorMulT achieves 83.4\%, surpassing the next best, i.e., MTFN, at 82.5\%. Similarly, for the F1 score, CorMulT's 83.5\% markedly eclipsed others, showcasing its robustness for binary sentiment classification.
Besides, CorMulT excels with a score of 0.691 in Corr(Correlation), substantially ahead of MLF-DNN's 0.675, underscoring its adeptness in comprehending multimodal data intricacies. Regarding Acc@3, CorMulT scores 68.0\%, maintaining its strong performance but lower than MLF-DNN's 69.0. By examining the results, we observe that our model performs well in distinguishing between positive and negative sentiments. However, seemingly due to the perturbations introduced during training, it tends to classify more ambiguous emotions as neutral (as shown in our results that there are more non-neutral samples classified as neutral). This contributes to a slightly lower performance in the triple class classification task. Nevertheless, CorMulT remains among the top-performing models in this aspect compared to others.

In sum, CorMulT model manifests significant superiority in multimodal sentiment analysis in terms of accuracy, F1 scores, and correlation.


\begin{table}[h]
    \centering
    \caption{Comparison of CorMulT with multiple baseline models for multimodal sentiment analysis on CMU\_ MOSEI dataset.}
    \begin{tabular}{llccccc}
        \toprule
        Model & Modality & Acc@7 & Acc@2 & F1 & Corr \\
        \midrule
        EF-LSTM(2018)\cite{williams-etal-2018-recognizing} & V+A+T & 43.8 & 75.2 & 75.3 & 0.609 \\
        LMF(2018)\cite{LMF} & V+A+T & 48.1 & 82.1 & 82 & 0.677 \\
        TFN(2017)\cite{TFN} & V+A+T & 50.2 & 82.5 & 82.1 & 0.701 \\
        MFM(2023)\cite{Li2023MultimodalFM} & V+A+T & 50.3 & 82.4 & 82.3 & 0.717 \\
        CM-Bert(2020)\cite{CM-Bert} & V+T & 50.5 & 83.6 & 83.5 & 0.747 \\
        MISA(2020)\cite{Hazarika1} & V+A+T & 50.7 & 83.9 & 83.8 & 0.748 \\
        BBFN(2021)\cite{Han1} & V+A+T & 50.8 & 82.2 & 82.2 & 0.761 \\
        MULT(2019)\cite{MULT} & V+A+T & 51.1 & 83.3 & 83.5 & 0.758 \\
        ICCN(2019)\cite{ICCN} & V+A+T & 51.2 & 82.2 & 82.4 & 0.713 \\
        MMLatch(2022)\cite{Paraskevopoulos1} & V+A+T & 52.1 & 83.3 & 83.9 & 0.754 \\
        TMSON(2024)\cite{TMSON} & V+A+T & 55.6 & 86.4 & 86.2 & 0.766 \\
        DVMI(2024)\cite{DVMI} & V+A+T & - & 86.6 & 86.4 & 0.773 \\
        ABSA-TLRF(2024)\cite{ABSA-TLRF} & V+A+T & 54.8 & 86.5 & 86.6 & -\\
        EMT(2023)\cite{EMT} & V+A+T & 54.5 & 86.0 & 86.0 & 0.774\\
        MMIN(2024)\cite{MMIN} & V+A+T & - & 85.9 & 85.8 & 0.761 \\
        \textbf{CorMulT} & V+A+T & \textbf{55.9} & \textbf{86.8} & \textbf{86.7} & \textbf{0.779} \\
        \bottomrule
    \end{tabular}
    \label{tab:comparison}
\end{table}

\begin{table}[h]
    \centering
    \caption{Comparison of CorMulT with multiple baseline models for multimodal sentiment analysis on CH-SIMS dataset.}
    \begin{tabular}{llcccc}
        \toprule
        Model & Modality & Acc@2 & Acc@3 & F1 & Corr \\
        \midrule
        EF-LSTM(2018)\cite{williams-etal-2018-recognizing} & V+A+T & 69.4 & 54.3 & 81.9 & 0.055 \\
        LF-DNN(2013)\cite{socher-etal-2013-recursive} & V+A+T & 79.0 & 64.3 & 80.2 & 0.555 \\
        TFN(2017)\cite{TFN} & V+A+T & 80.7 & 65.1 & 81.6 & 0.591 \\
        LMF(2018)\cite{LMF} & V+A+T & 79.2 & 64.7 & 81.6 & 0.576 \\
        MFM(2018)\cite{MFM} & V+A+T & 77.9 & 65.7 & 79.9 & 0.582 \\
        Graph-MFN(2018)\cite{Graph-MFN} & V+A+T & 78.8 & 65.7 & 78.2 & 0.578 \\
        MulT(2019)\cite{MULT} & V+A+T & 78.6 & 64.8 & 79.7 & 0.564 \\
        MLF-DNN(2020)\cite{CH-SIMS} & V+A+T & 82.3 & \textbf{69.0} & 82.5 & 0.675 \\
        MTFN(2020)\cite{CH-SIMS} & V+A+T & 82.5 & 68.8 & 82.6 & 0.670 \\
        Self-MM(2021)\cite{Self-MM} & V+A+T & 80.0 & 65.5 & 80.4 & 0.592 \\
        DVMI(2024)\cite{DVMI} & V+A+T & 81.2 & - & 81.3 & 0.599 \\
        M\textsuperscript{3}SA(2024)\cite{M3SA} & V+A+T & 81.6 & 68.1 & 81.5 & 0.614 \\
        EMT(2023)\cite{EMT} & V+A+T & 80.1 & 67.4 & 80.1 & 0.623 \\
        \textbf{CorMulT} & V+A+T & \textbf{83.4} & 68.0 & \textbf{83.5} & \textbf{0.691} \\
        \bottomrule
    \end{tabular}
    \label{tab:ch_sims_comparison}
\end{table}

\subsection{Visualization Analysis}
\begin{figure}[t]
  \centering
  \includegraphics[width=0.95\linewidth]{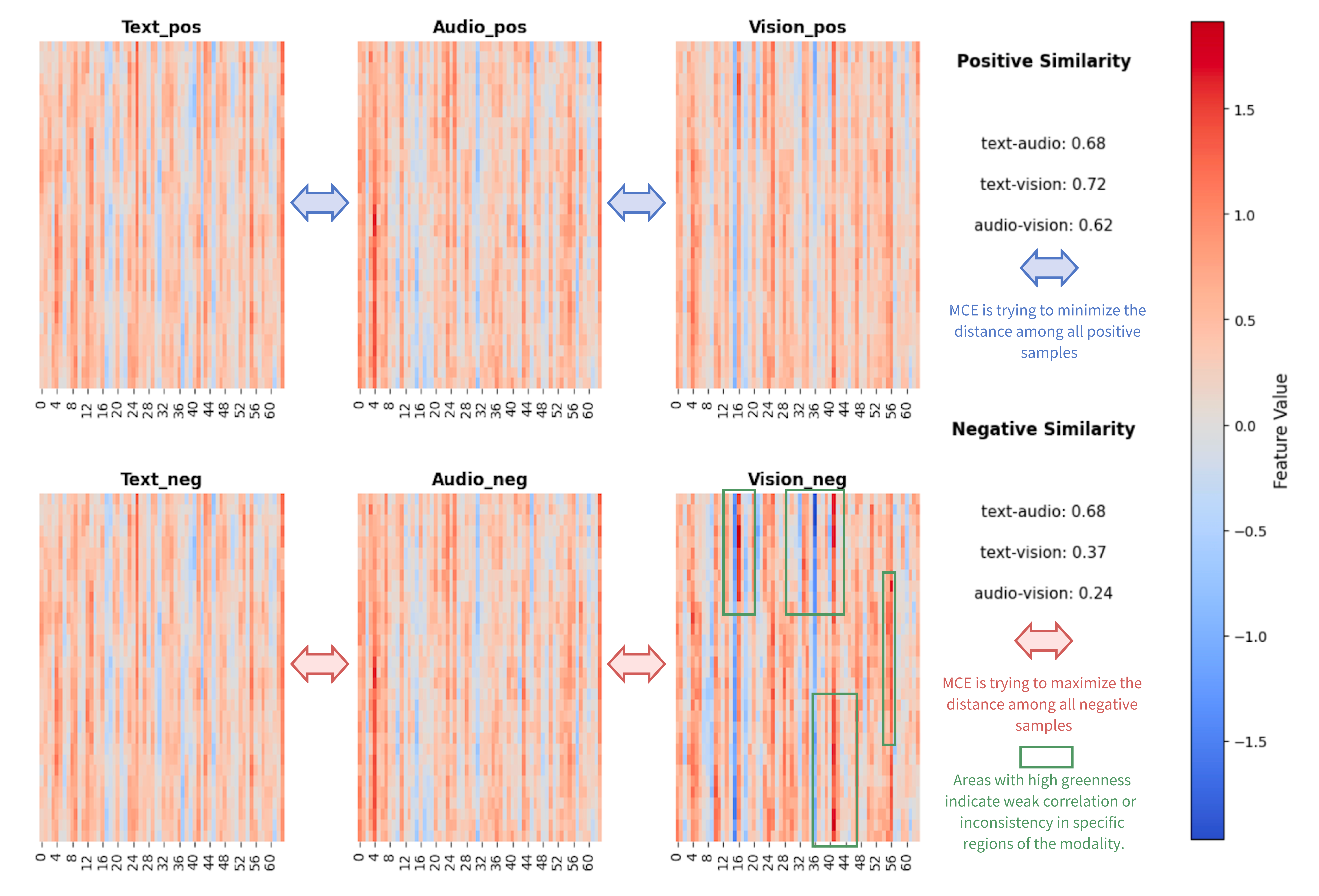}
  \caption{Visualization of the final-layer MCE features for three positive samples (top row) and three negatively perturbed samples (bottom row). The color bar on the right indicates the feature values, with redder tones signifying higher activation. On the far right, the corresponding pairwise cross-modal similarities are shown, illustrating strong positive correlations in the top row and reduced Vision correlations in the negative samples.}
  \label{fig:viz_posneg}
\end{figure}

To better demonstrate how the proposed Modality Correlation Evaluation (MCE) module captures and distinguishes modality relationships, we visualize in Fig.~\ref{fig:viz_posneg} the final-layer representations (after MCE) for three positive samples and three negative samples. Each subfigure displays a heatmap for one modality (Text, Audio, or Vision), where the vertical axis corresponds to time and the horizontal axis represents the feature dimension. The color intensity indicates the learned feature values; redder tones reflect higher feature values, whereas bluer tones indicate lower feature values. In this example, we employ our perturbation ``strategy 3'' by injecting Gaussian noise into the Vision modality to create negative samples. 

\textbf{Positive vs.\ Negative Samples.}
In the top row of Fig.~\ref{fig:viz_posneg}, we show the learned representations for the three modalities when they are highly correlated (positive samples), while the bottom row demonstrates the outcome for negatively perturbed samples (negative samples). As shown in the figure, the Vision representations in the negative case appear more misaligned and deviate substantially from the other two modalities (Text and Audio).  

On the right side of Fig.~\ref{fig:viz_posneg}, we also include quantitative similarity scores measured by cosine similarity. One can observe that for \emph{positive} samples, the cross-modal similarities (Text--Audio, Text--Vision, and Audio--Vision) remain consistently high (around $0.6$--$0.7$). By contrast, in the \emph{negative} samples, only the Text--Audio similarity stays at a relatively higher value (around $0.6$--$0.7$), while both Text--Vision and Audio--Vision drop to $\approx 0.4$ or lower. This indicates that the model correctly identifies the perturbation in the Vision modality and registers much lower correlation between Vision and the other two modalities.

\begin{figure}[t]
  \centering
  \includegraphics[width=0.95\linewidth]{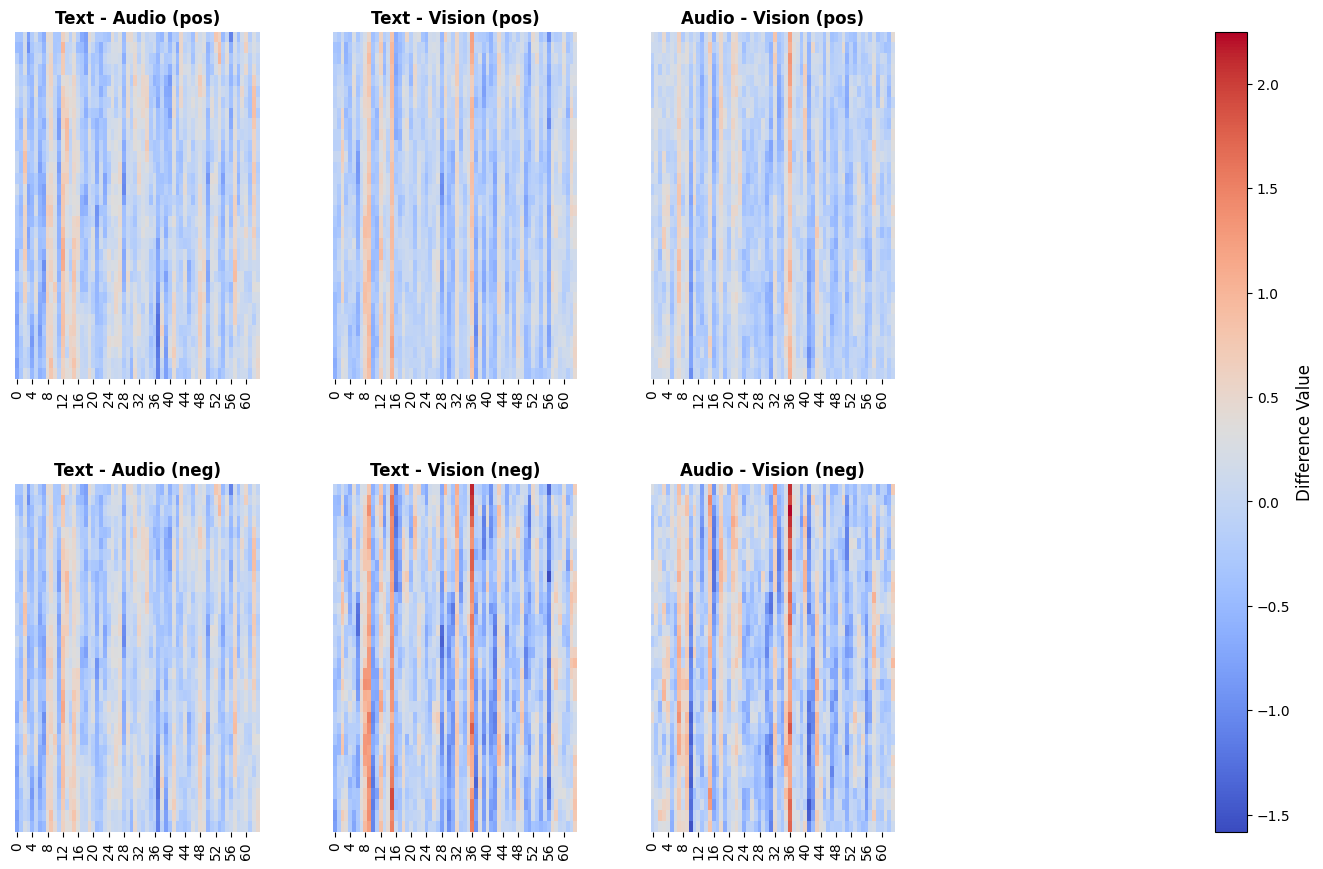}
  \caption{Heatmaps of pairwise feature \emph{differences} between modalities, with the top row corresponding to positive samples and the bottom row to negative samples. Larger red/blue patches correspond to bigger differences. Notice that in the negative samples, the difference maps (especially for Vision) exhibit much larger intensities.}
  \label{fig:viz_diff}
\end{figure}

\textbf{Difference Maps.}
To further highlight how perturbations affect modality relationships, we plot in Fig.~\ref{fig:viz_diff} the pairwise \emph{difference} heatmaps between modalities. For instance, \emph{Text--Audio (pos)} displays the difference between the Text and Audio features in a positive sample, while \emph{Text--Vision (neg)} shows the difference for a negative sample where Vision has been perturbed. As the color scale indicates, deeper red/blue patches imply larger absolute differences. We observe that for \emph{positive} samples, the differences between any two modalities remain relatively small, reflecting their higher correlation. In contrast, the \emph{negative} examples exhibit markedly higher difference values in \emph{Text--Vision} and \emph{Audio--Vision}, reinforcing the fact that these modalities have become misaligned due to the Gaussian noise applied to Vision.

These visualizations confirm that our proposed MCE module indeed captures modality correlations effectively. When modal inputs are properly aligned and correlated (positive samples), the learned representations show relatively small differences and high cross-modal similarity. However, once a modality (Vision in this case) is perturbed or not correlated as negative samples, the MCE outputs reflect comparably lower similarity with the unperturbed modalities. This demonstrates the module’s capability to discern correlated vs.\ uncorrelated modalities, thereby enhancing the overall transformer architecture’s ability to learn robust multimodal representations.

\subsection{Weak Modality Correlation Analysis}
This section provides targeted experiments to evaluate model performance on four types of weak correlations and compares results with both strongly correlated and baseline cases, i.e., voice discrepancy, content discrepancy, alignment discrepancy, and clarity discrepancy.

These discrepancies are tied to three negative-sample perturbation strategies in our contrastive learning module:
\textit{Strategy A} randomly replaces one modality with another sample (e.g., generating voice/content discrepancy), 
\textit{Strategy B} shifts the temporal alignment among modalities (alignment discrepancy), 
and \textit{Strategy C} injects noise or random corruption (clarity discrepancy). Specifically, these scenarios are simulated with the negative sample generation strategies by:
\begin{itemize}
    \item \textbf{Voice Discrepancy}: Audio content is replaced or shifted, leading to mismatched semantics with the text/visual streams.
    \item \textbf{Content Discrepancy}: Text or visual frames are replaced by other samples, inducing semantic inconsistencies.
    \item \textbf{Alignment Discrepancy}: Audio, visual, or textual elements are temporally offset, introducing misalignment among modalities.
    \item \textbf{Clarity Discrepancy}: Noise or blurring is added to audio/visual modalities, degrading signal quality.
\end{itemize}

\paragraph{Comparison with Strongly Correlated Data}
Table~\ref{tab:wc_strong_compare} first shows that on the \emph{strongly correlated} portion of CMU\_MOSEI, our CorMulT achieves $55.9\%$ on Acc@7 and $86.7\%$ on F1. Once the dataset is disturbed with random weak-correlation perturbations, the performance drops to $51.7\%$ (Acc@7) and $82.3\%$ (F1), but still substantially outperforms the baseline MULT~\cite{MULT} under the same perturbations (Acc@7: $47.5\%$, F1: $76.3\%$). This validates that our modal-correlation contrastive learning framework effectively mitigates the adverse impacts of weak correlation.

\begin{table}[!h]
\centering
\caption{CorMulT vs.\ MULT on strongly vs.\ weakly correlated data (overall random perturbations).}
\label{tab:wc_strong_compare}
\begin{tabular}{lcccc}
\toprule
\textbf{Method} & \textbf{Data Type} & \textbf{Acc@7} & \textbf{F1} & \textbf{Corr} \\
\midrule
CorMulT & Strongly Corr.\  & 55.9 & 86.7 & 0.779 \\
CorMulT & Weakly Corr.\    & 51.7 & 82.3 & 0.735 \\
\midrule
MULT    & Weakly Corr.\    & 47.5 & 76.3 & 0.635 \\
\bottomrule
\end{tabular}
\end{table}

\paragraph{Analysis of Four Weak-Correlation Types}
To systematically evaluate our model’s robustness against different forms of weak modality correlation, we design four controlled perturbation scenarios, each reflected by a distinct negative-sample generation strategy: 
(\emph{Strategy~A}) random replacement for \textbf{voice/content discrepancy}, 
(\emph{Strategy~B}) temporal shifting for \textbf{alignment discrepancy}, 
and (\emph{Strategy~C}) noise corruption for \textbf{clarity discrepancy}. 
Table~\ref{tab:wc_comparison} reports the comparison between our \textit{CorMulT} and a standard multimodal transformer \textit{MULT} under each discrepancy type. We observe:

\begin{itemize}
    \item \textbf{Content/Voice Discrepancy (Strategy A).} 
    By replacing the text or visual modality (content discrepancy) or substituting the modality  (voice discrepancy), we induce comparable semantic conflicts across modalities. As expected, both models suffer a notable performance drop. Nonetheless, \textit{CorMulT} still benefits from partially correlated cues in the unaffected modalities, retaining mid-70\% F1 and consistently outperforming \textit{MULT} in accuracy, F1, and correlation.

    \item \textbf{Alignment Discrepancy (Strategy B).}
    Time misalignment is introduced by shifting the timeline of one modality. \textit{CorMulT}’s correlation-aware design and contrastive objectives help mitigate the negative effects of misalignment more effectively, preserving higher accuracy and stronger correlation compared to \textit{MULT}.

    \item \textbf{Clarity Discrepancy (Strategy C).}
    With Strategy C, we degrade one modality’s correlation and quality via noise addition or blurring. Despite the corruption, \textit{CorMulT} exhibits relatively stable performance, suggesting robustness to signal noise. This contrasts with the more pronounced degradation observed in \textit{MULT}, indicating that \textit{CorMulT}’s contrastive approach better preserves essential features under noise perturbation.
\end{itemize}

\begin{table}[!ht]
\centering
\caption{Comparison of \textit{CorMulT} and \textit{MULT} under different weak-correlation types. Each discrepancy is simulated by a distinct negative-sample generation strategy (A/B/C).}
\label{tab:wc_comparison}
\begin{tabular}{llcccc}
\toprule
\textbf{Discrepancy} & \textbf{Method} & \textbf{Acc@7 (\%)} & \textbf{Acc@2 (\%)} & \textbf{F1 (\%)} & \textbf{Corr} \\
\midrule
\multirow{2}{*}{\makebox[5em]{Content (A)}} 
    & \textit{CorMulT} & 44.2 & 74.4 & 74.4 & 0.639 \\
    & \textit{MULT}   & 41.5 & 70.2 & 70.5 & 0.598 \\
\midrule
\multirow{2}{*}{\makebox[5em]{Voice (A)}} 
    & \textit{CorMulT} & 47.6 & 76.8 & 76.8 & 0.657 \\
    & \textit{MULT}   & 45.1 & 73.8 & 73.6 & 0.623 \\
\midrule
\multirow{2}{*}{\makebox[5em]{Alignment (B)}} 
    & \textit{CorMulT} & 53.3 & 85.2 & 85.2 & 0.752 \\
    & \textit{MULT}   & 48.7 & 78.5 & 78.8 & 0.690 \\
\midrule
\multirow{2}{*}{\makebox[5em]{Clarity (C)}} 
    & \textit{CorMulT} & 53.6 & 84.0 & 84.9 & 0.749 \\
    & \textit{MULT}   & 48.2 & 77.8 & 78.1 & 0.682 \\
\bottomrule
\end{tabular}
\end{table}

\vspace{0.3em}
\noindent
\textbf{Conclusion on Weak-Correlation Analysis.}
Our targeted experiments demonstrate that \textit{CorMulT} effectively mitigates the adverse effects of mismatched or corrupted modalities. Across voice, content, alignment, and clarity discrepancies, it preserves both accuracy and correlation better than baseline methods, confirming the strength of its correlation-aware architecture and contrastive learning in real-world multimodal scenarios with imperfectly aligned or noisy streams.

\subsection{Ablation Studies}
We conducted two sets of ablation studies to validate the influence of MCE multimodal feature extraction module and the number of modalities used for pre-training CorMulT model.

For modality-specific feature extraction, we applied tailored approaches to address the distinct characteristics of three different data types. To gauge the effectiveness of these approaches, we specifically examined the transformation of the Mel spectrum matrix for audio data and the application of ResNet50 for image feature mapping. Table~\ref{tab:modality_ablation} quantitatively evaluates the effectiveness of different feature extraction strategies in the MCE module across various modalities. The strategies include the use of Mel spectrum transformation for audio data, ResNet50 for image data, both methods combined, and a baseline with no specialized feature extraction methods. Metrics assessed include loss values for audio (\(Loss_{A}\)), text (\(Loss_{T}\)), and visual (\(Loss_{V}\)) data, along with the average loss (\(Loss_{ave}\)) and epoch time (\(t_{epoch}\)).
The results demonstrate the efficacy of modality-specific feature extraction. Tailored approaches such as Mel spectrum for audio and ResNet50 for visual data significantly reduce the losses across respective modalities, with the combined approach yielding the lowest average loss (\(Loss_{ave}\) = 0.168) and shortest training time (41.02 seconds per epoch). This verifies the importance of employing specific feature extraction methods for each modality.

\begin{table}[h]
    \centering
    \caption{The results of CorMulT while using different modality-specific feature extraction methods.}
    \begin{tabular}{lccccc}
        \toprule
        Strategy & \(Loss_{A}\) & \(Loss_{T}\) & \(Loss_{V}\) & \(Loss_{ave}\) & \(t_{epoch}\) \\
        \midrule
        None        & 0.251 & 0.219 & 0.474 & 0.315 & 178.24 \\
        Mel(Audio)  & 0.191 & 0.169 & 0.449 & 0.270 & 175.43 \\
        Res(Vision) & 0.149 & 0.143 & 0.232 & 0.174 & 43.14 \\
        Full        & \textbf{0.136} & \textbf{0.142} & \textbf{0.225} & \textbf{0.168} & \textbf{41.02} \\
        \bottomrule
    \end{tabular} 
    \label{tab:feature_ablation}
\end{table}

We further evaluated CorMulT under different MCE pre-training configurations: no pre-training (None), audio and text only (A+T), visual and text only (V+T), and all three modalities (V+A+T). As shown in Table~\ref{tab:modality_ablation}, even without MCE pre-training, CorMulT still achieves competitive performance due to its robust MulT-based backbone. Nevertheless, incorporating additional pre-trained modalities within MCE steadily improves the results because the model acquires richer cross-modal correlation knowledge. In particular, expanding from a single pre-trained modality to two or three yields increasingly better synergy with CorMulT’s crossmodal attention layers, ultimately boosting sentiment analysis accuracy. These findings confirm that while the underlying MulT structure provides a strong starting point, the model's performance benefits significantly from our correlation-aware design and modality pre-training.

\begin{table}[h]
    \centering
    \caption{The results of CorMulT while using different modalities for pre-training.}
    \begin{tabular}{lccccc}
    \hline
        \toprule
        Modalities & Acc@7 & Acc@2 & F1 & Corr \\
        \midrule
        None & 50.9 & 82.9 & 83.4 & 0.752 \\
        A+T & 51.2 & 83.2 & 83.8 & 0.762 \\
        V+T & 52.3 & 86.2 & 86.1 & 0.778 \\
        V+A+T & \textbf{52.8} & \textbf{86.5} & \textbf{86.7} & \textbf{0.779} \\
        \bottomrule
    \end{tabular}
    \label{tab:modality_ablation}
\end{table}

\section{Conclusion}\label{sec:conclusion}
In this work, we proposed the two-stage semi-supervised model CorMulT which designs a modality correlation contrastive learning module to efficiently learn modality correlation coefficients between different modalities, and fuses the learned correlation coefficients with modality representations to make accurate sentiment prediction. According to the experimental evaluation, CorMulT obviously outperform existing multimodal sentiment analysis methods.
In addition, the ablation studies also 
demonstrate the significance of mutual information between modalities in improving the performance of sentiment analysis models. With its excellent stability and high performance, CorMulT model achieves effective utilization of intermodal correlations and provides a new solution to multimodal sentiment analysis.
As for future work, it is promising to further explore the dynamics of intermodal correlations for better multimodal sentiment analsyis.

\section*{Acknowledgments}
This work was supported in part by the Open Research Projects of Zhejiang Lab (No. 2021KH0AB04), the National Natural Science Foundation of China (No. 62202336), and the Fundamental Research Funds for the Central Universities (No. 2024-4-YB-03).

We would also like to thank Xinhang Yuan from Washington University in St. Louis (WashU) for her significant contributions during the revision phase of this work. While she is not listed as a co-author, she proposed an innovative training strategy incorporating weakly correlated multimodal data, which addressed and improved the model’s robustness. Xinhang helped with the design and execution of new experiments, extended our evaluations to additional datasets, and developed visualization tools to enhance the manuscript’s clarity.

\bibliographystyle{IEEEtran}
\bibliography{citations}











\newpage
\begin{IEEEbiography}[{\includegraphics[width=1in,height=1.25in,clip,keepaspectratio]{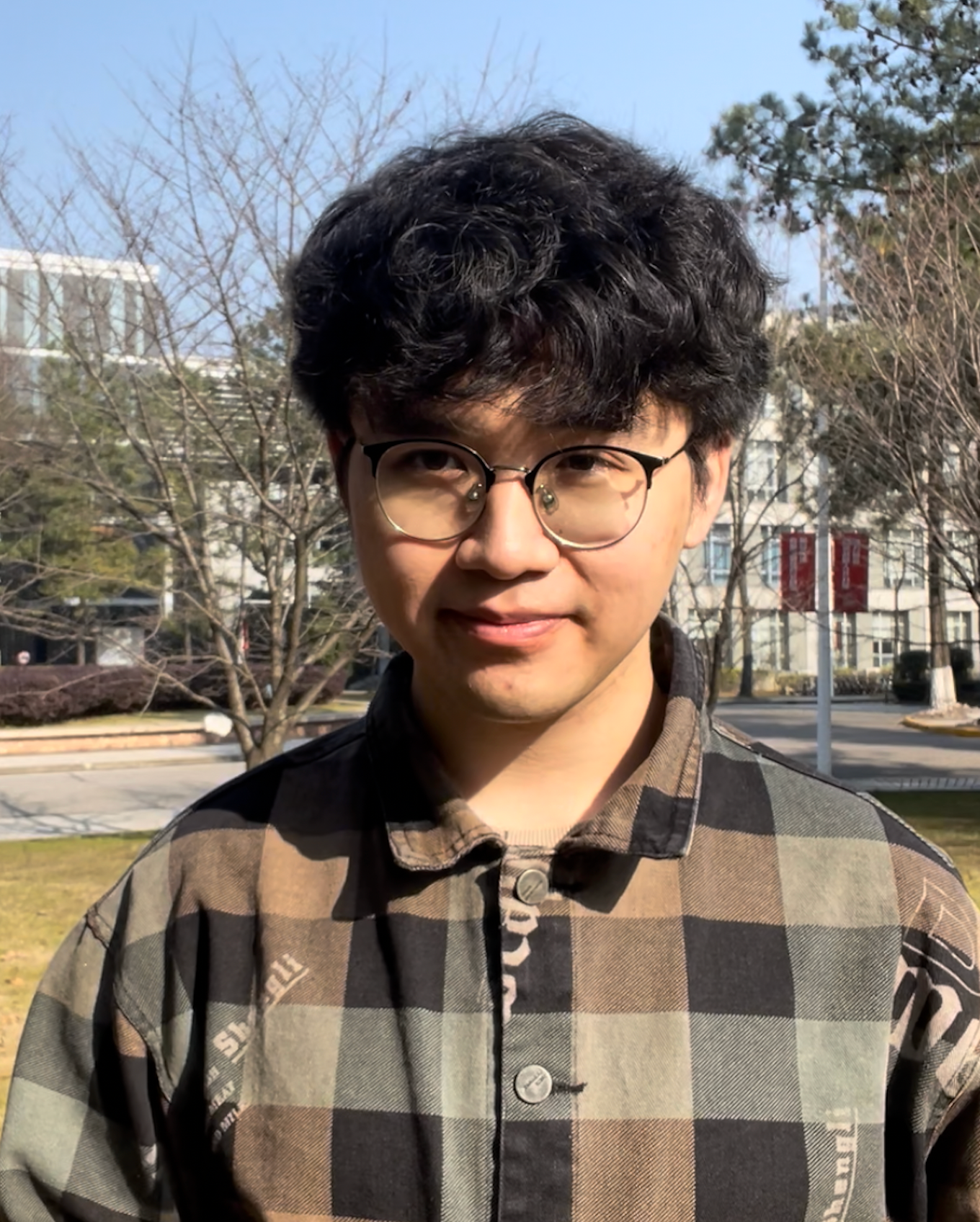}}]{Yangmin Li}
received his B.Eng. degree in Computer Science from Tongji University, Shanghai, China (2019–2023), and his Master’s degree from Carnegie Mellon University (CMU), where he focused on Distributed Systems and Cloud Computing. He is currently working as an AI Platform Software Engineer at NVIDIA, where he focuses on advancing AI technologies in autonomous driving and related domains. His research interests include natural language processing, multimodal machine learning, and high-performance optimization algorithms, underpinned by a strong foundation in distributed systems and cloud computing. He is a member of IEEE and ACM.
\end{IEEEbiography}

\vspace{-100mm}
\begin{IEEEbiography}[{\includegraphics[width=1in,height=1.25in,clip,keepaspectratio]{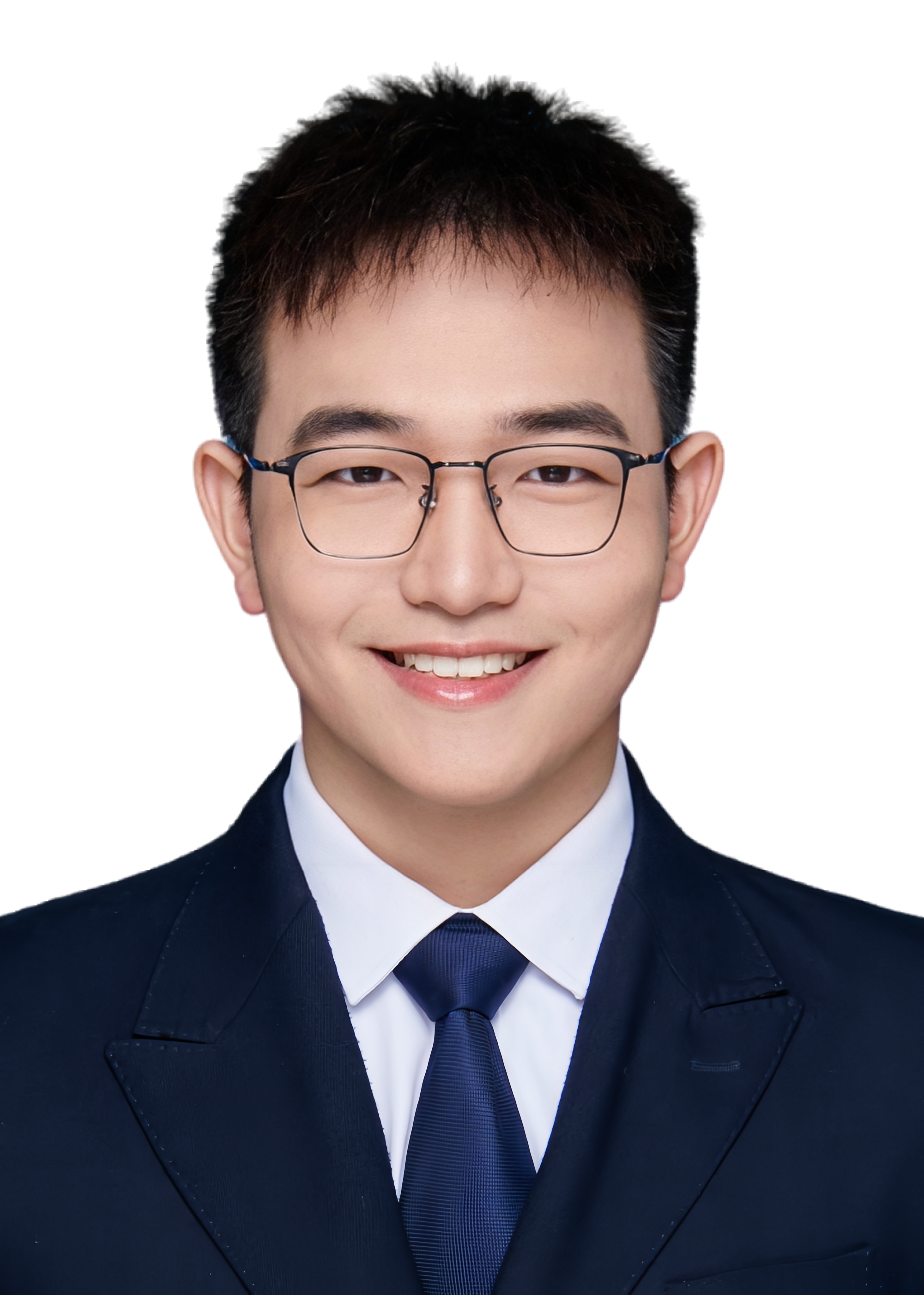}}]{Ruiqi Zhu}
received a bachelor degree in computer science and technology from Tongji University in 2024. He has worked as a Database Product Manager at Transwarp and plans to pursue a Master's degree in Hong Kong or Singapore. He has participated in national and international mathematical modeling competitions, achieving commendable results. His research interests include VQA, multimodal sentiment analysis and cloud computing.
\end{IEEEbiography}
\vspace{-100mm}
\begin{IEEEbiography}[{\includegraphics[width=1in,height=1.25in,clip,keepaspectratio]{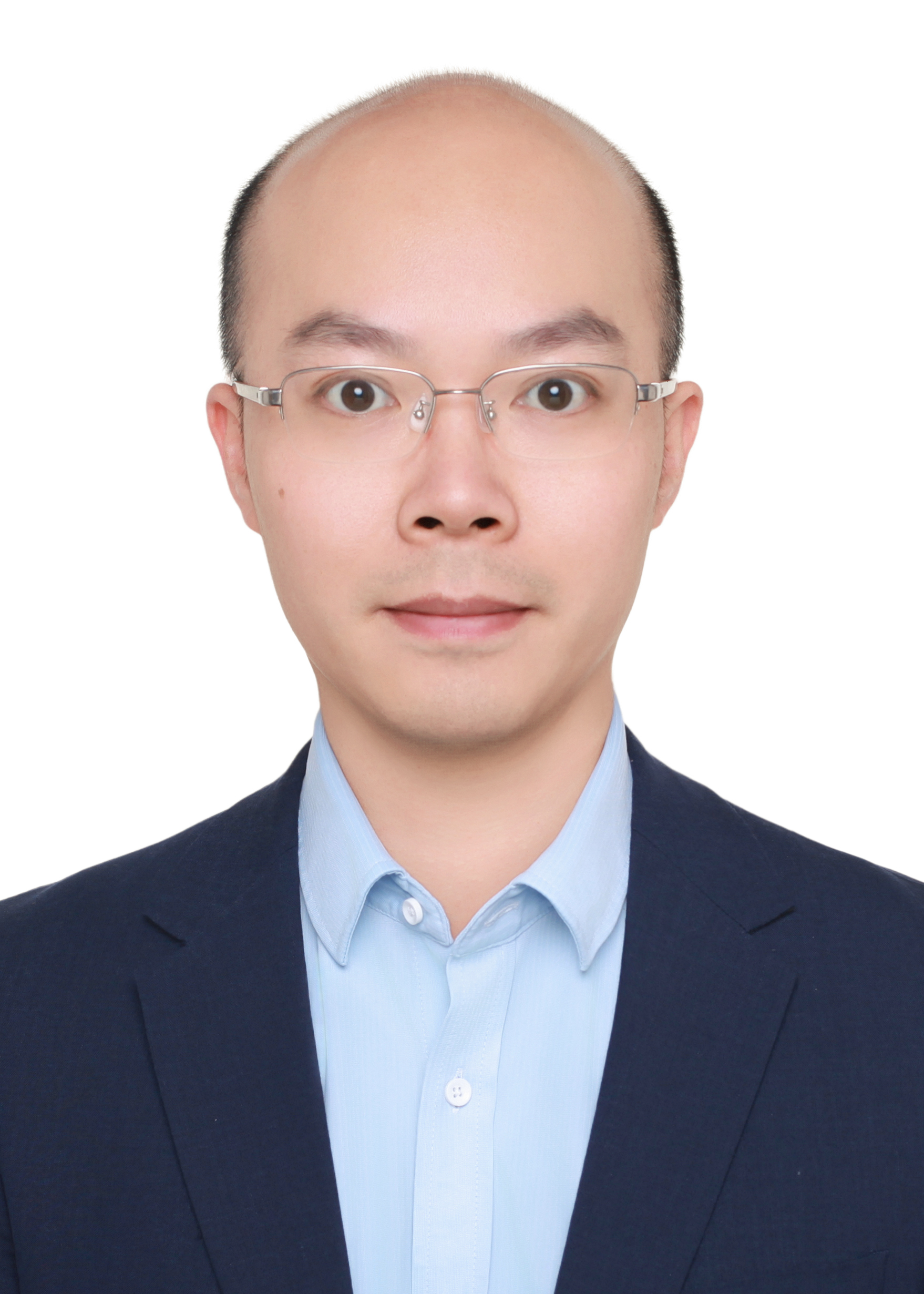}}]{Wengen Li}
received the B.Eng. degree and Ph.D. degree in Computer Science from Tongji University, Shanghai, China, in 2011 and 2017, respectively. In addition, he received a dual Ph.D. degree in Computer Science from the Hong Kong Polytechnic University in 2018. He is currently an associate professor of the School of Computer Science and Technology at Tongji University. His research interests include multi-modal data analytics, and spatio-temporal data analytics for urban computing and ocean computing. He is a member of China Computer Federation (CCF), a member of IEEE, and a member of ACM.
\end{IEEEbiography}



\end{document}